%% file: root.tex
\documentclass{article}

\usepackage{microtype}
\usepackage{graphicx}
\usepackage{subcaption}
\usepackage{booktabs} 

\usepackage{natbib}
\renewcommand{\cite}{\citep}
\setcitestyle{authoryear} 

\usepackage[margin=1in]{geometry}

\usepackage{hyperref}

\usepackage{mathpazo}

\usepackage{amsmath}
\usepackage{amssymb}
\usepackage{mathtools}
\usepackage{amsthm}

\usepackage[capitalize,noabbrev]{cleveref}

\theoremstyle{plain}
\newtheorem{theorem}{Theorem}[section]

\theoremstyle{definition}
\newtheorem{definition}[theorem]{Definition}

\theoremstyle{remark}

\input{macros.tex}

\title{Path Signatures and Graph Neural Networks for Slow Earthquake Analysis: Better Together?}

\author{
    Hans Riess\thanks{Department of Electrical \& Computer Engineering, Duke University, Durham, North Carolina, USA.} \and
    Manolis Veveakis\thanks{Department of Civil \& Environmental Engineering, Duke University, Durham, North Carolina, USA.} \and
    Michael M. Zavlanos\thanks{Department of Mechanical Engineering \& Materials Science, Durham, North Carolina, USA.}
}

\date{}

\begin{document}

\maketitle

\vskip 0.3in

\begin{abstract}
    The path signature, having enjoyed recent success in the machine learning community, is a theoretically-driven method for engineering features from irregular paths. On the other hand, graph neural networks (GNN), neural architectures for processing data on graphs, excel on tasks with irregular domains, such as sensor networks. In this paper, we introduce a novel approach, Path Signature Graph Convolutional Neural Networks (PS-GCNN), integrating path signatures into graph convolutional neural networks (GCNN), and leveraging the strengths of both path signatures, for feature extraction, and GCNNs, for handling spatial interactions. We apply our method to analyze slow earthquake sequences, also called slow slip events (SSE), utilizing data from GPS timeseries, with a case study on a GPS sensor network on the east coast of New Zealand's north island. We also establish benchmarks for our method on simulated stochastic differential equations, which model similar reaction-diffusion phenomenon. Our methodology shows promise for future advancement in earthquake prediction and sensor network analysis.
\end{abstract}

\begin{figure}[t]
    \includegraphics[width=\linewidth]{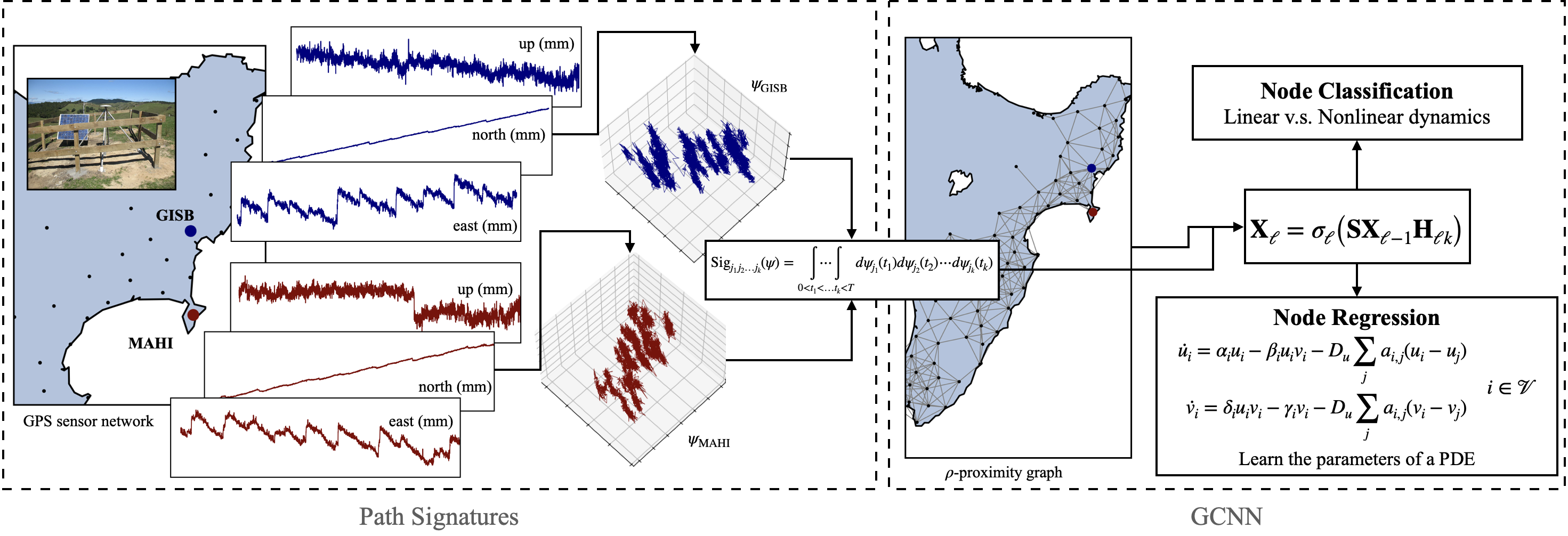}
    \caption{Our \textbf{PS-GCNN} pipeline consists of two modules: a Path Signatures module (left) and a Graph Convolutional Neural Network (GCNN) module (right). In our experiments with real data, GPS signals measuring the movement of the earth's crust are collected at stations (GISB and MAHI pictured), interpolated as $3$-dimensional paths, and transformed into node features. For a given radius $\rho\geq0$, the feature matrix $\bX$ is fed into a GCNN which is customized to perform either a node classification task or a node regression task.}
    \label{fig:summary}
\end{figure}

\section{Introduction}
\label{sec:intro}

The prediction of earthquakes has long been considered to be cumbersome due to its random and stochastic nature \cite{verejones2011}. The identification of frequently occurring %
\emph{slow slip events} (SSE), otherwise known as slow earthquakes \cite{obara2002}, at subduction interfaces has led to the premise that GPS time series measuring land surface displacement can recover the true nonlinear or even chaotic signal underpinning the occurrence of earthquakes \cite{poulet2013}. Since we currently have dense networks of such GPS sensors, the spatio-temporal analysis of the signals requires the development of space-sensitive reduction tools. In this work, we emphasize the slow earthquake sequences of the north island of New Zealand \cite{wallace2020}. This area features a dense network of GPS stations with displacement measurements recorded continuously for over a decade (see Fig.~\ref{fig:summary}). 

The north island of New Zealand sits at the intersection of three slow earthquake sequences (see  Figure 2 of \citet{wallace2020}) caused by the shallow subducting Hikurangi trench on the east, the deep signal under the Taupo volcanic area in the west, and the alpine fault system in the south. These three sequences have distinctly different nonlinear time-sequences (i.e.~exhibiting positive maximal Lyapunov exponent in a dynamic analysis, see \citet{abarbanel1992}) that can interfere with the GPS signals of the stations across the island. An open question in processing the signals from the GPS network is how to identify the origins of the signal with respect the three slow earthquake sequences and correlate the stations according to that origin. To this end we have focused on 80 GPS stations at the east coast of the island (see Fig.~\ref{fig:summary}) which exhibit a sufficiently complicated signal interference and seek to develop an approach that can identify the nonlinear features of a signal and classify them in different locations.

\begin{figure}[h]
    \centering
    \includegraphics[width=0.5\linewidth]{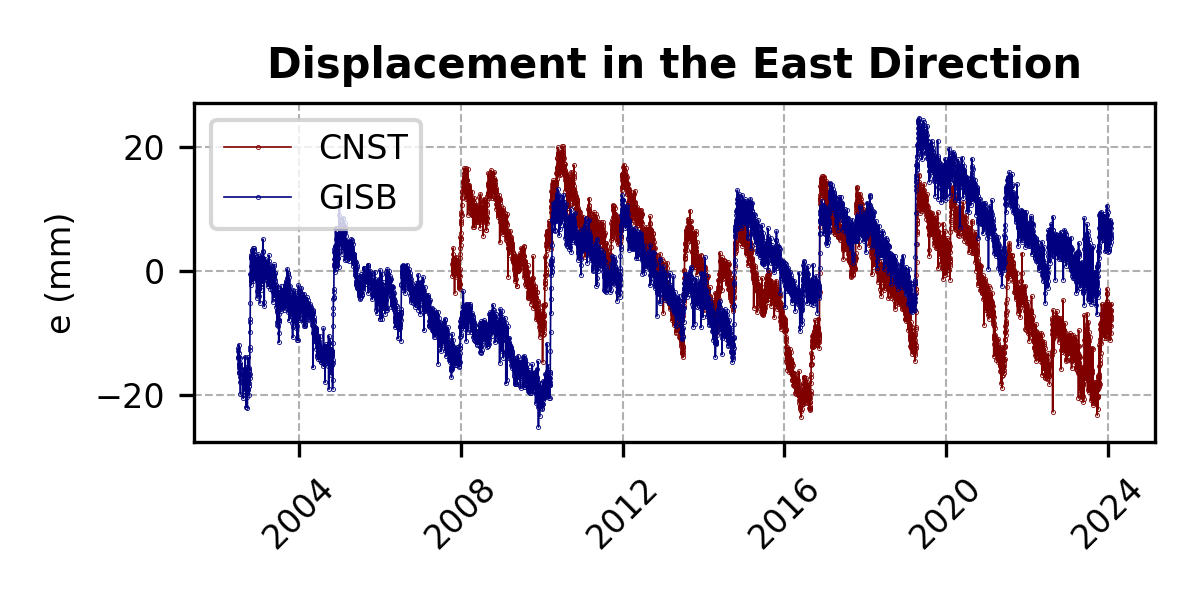}
    \caption{Lead-lag behavior aparent in the east displacement signal for nearby stations GISB and CNST.}
    \label{fig:lead-lag}
\end{figure}

In that respect, path signatures extract features from paths, maps from an interval $[0,T]$ into a Euclidean space, that encode many of their key geometric characteristics, including lead-lag behavior \cite{flint2016}, as exhibited between nearby GPS signals (see Fig.~\ref{fig:lead-lag}). Path signatures have been compared to fourier transforms \cite{kidger2019}, instead extracting information about area \cite{hambly2010} and order \cite{gyurko2013}, as opposed to classical frequency. Due to their general definition---iterated integrals--- path signatures can be defined in many general settings, including continuous paths of bounded variation in the original definition \cite{chen1957}, rough paths \cite{lyons1998}, piece-wise differentiable paths \cite{pfeffer2019}, as well paths in non-Euclidean spaces such as Lie groups \cite{lee2020}. In data-driven applications such as ours, however, paths come as streams, sequences of data, or, equivalently, samples from multivariate time series, resulting in piece-wise linear paths via interpolating the data.

While the data to path signature pipeline has made made path signatures an increasingly attractive tool in machine learning \cite{chevyrev2016,lyons2022,kidger2019}, to the best of our knowledge, path signatures have not been previously utilized as node features in graph neural networks (GNNs), neural architectures for processing data on graphs \cite{ruiz2021,wu2020}. Thus, we believe analysis of slow slip events with GPS time series is fertile ground for applying the path signature method in the graph setting, as the sensors quite literally trace paths as a result of land surface displacement.

To this end, our first contribution is a pipeline which we call \emph{Path Signature Graph Convolutional Neural Network} (PS-GCNN). We apply this pipeline to the GPS sensor network (as well as simulated sensor networks) on node-level tasks. One one hand, we describe a data-driven approach to make inferences about the dynamics of slow slip events through a semi-supervised node classification task. On the other hand, we introduce a general methodology for addressing a wider range of learning tasks for data presented as (or transformed into) multivariate time series on graphs.

When a physical model of a system is unknown, path signatures can be used to engineer features that represent observables of the system. We explore the setting where the form of a physical model is known except for a few scalar parameters, as studied in pioneering work in physics-informed machine learning \cite{raissi2018}.
Trajectories of partial differential equations (PDEs) supply a class of examples of spatial-temporal time series obtained by fixing the spatial variable and numerically solving the resulting ordinary differential equation (ODE). Because the GPS signals are in fact stochastic proceses, which were an original motivation for path signatures \cite{lyons1998}, we direct our attention towards stochastic PDEs, in particular, the stochastic reaction-diffusion equation
\begin{align}
\begin{aligned}
    \partial_t u(\mathbf{y}, t) &= R_u(u, v) + D_u \nabla^2 u + \sigma_u \eta_u(\mathbf{y}, t) \\
    \partial_t v(\mathbf{y}, t) &= R_v(u, v) + D_v \nabla^2 v + \sigma_v \eta_v(\mathbf{y}, t) \label{eq:reaction-diffusion}
\end{aligned}
\end{align}
where $u$ and $v$ are spatio-temporal signals, $R_u$ and $R_v$ describes a reaction between $u$ and $v$, $D_u$ and $D_v$ are coefficients of diffusivity, and $\sigma_u$ and $\sigma_v$ are the amplitudes of noise of a spatially-extended Weiner process $\eta_u(\by,t)$ and $\eta_v(\by,t)$. We remark, perhaps as a soft motivation for studying these equations, that trajectories of stochastic reaction-diffusion equations bear resemblance to GPS signals received during the occurrence of slow slip events (see Fig.~\ref{fig:lead-lag} and Fig.~\ref{fig:simulated-data}). Using simulated reaction-diffusion equations, we learn the parameters of the reaction terms, $R_u$ and $R_v$.

\section{Methods}
\label{sec:method}

 Critical challenges for tasks, such as forecasting, anomaly detection, and time series in this graph setting are graph engineering (if the graph is not specified) and feature engineering, which include node and edge features. Feature engineering is the task of representing complex data by (hopefully) informative \emph{features} \cite{dong2018}. In graph machine learning, feature engineering is particularly relevant because node features are often not provided in the raw data. Feature engineering is also paramount in time series analysis because raw series features are high-dimensional, leading to the curse-of-dimensionality. State-of-the-art feature engineering from multivariate time series relies on statistical summaries \cite{fulcher2018}. In our ablation study (Section \ref{sec:experiments}), we use an off-the-shelf feature generator \cite{christ2018} which extracts statistical summaries from multivariate timeseries data. A competing approach is a so-called backbone, a neural network module at the beginning of a machine learning pipeline trained to extract features \cite{elharrouss2022,mishra2020}. We, instead, propose using path signatures to generate node features to feed into a graph neural network architecture.

\subsection{Path signatures}
\label{sec:signatures}

A path is simply a map $\psi: [0,T] \to \R^n$ from an interval of time to Euclidean space. We write paths as tuples of time series $\psi = (\psi_1, \psi_2, \dots, \psi_n)$ such that each \emph{component} $\psi_i: [0,T] \to \R$ is a function assumed continuous and piece-wise smooth. We say $n$ is the number of \emph{dimensions} of the path. Frequently, it is informative to incorporate time as a separate dimension of a path. We say $\hat{\psi}: [0,T] \to \R^{n+1}$ is the \emph{augmented path} given by $\hat{\psi}(t) = \left(t, \psi_1(t), \psi_2(t), \dots, \psi_n(t) \right)$.

\begin{definition}[Path Signature]
    Suppose $\psi: [0,T] \to \R^n$ is a path. The iterated integral with respect to the multi-index $I = (i_1,i_2,\dots,i_k) \in \{1,2,\dots,n\}^k$ is
    \begin{align}
        \Sig_I(\psi) =  \underset{0 < t_1 < \dots t_k < T}{\int \cdots \int} d\psi_{j_1}({t_1}) d\psi_{j_2}({t_2}) \cdots d\psi_{j_k}({t_k})
    \end{align}
    where (whenever defined) $d\psi_j(t) = \psi_j'(t) dt$. The \emph{path signature of} $\psi$ is the series of coefficients $\Sig(\psi) = \bigl( \Sig_{I}(\psi) \bigr)_{I}$ indexed over all multi-indices.
\end{definition}


As first observed by \cite{chen1957}, the path signature can be viewed as a mapping $\Sig: \Path_{0,T} \left( \R^n \right) \to T((\R^n))$ where $T((\R^n)) = \bigoplus_{k=1}^\infty \left( \R^n \right)^{\otimes k}$ is the so-called ``tensor algebra'' indexing the coefficients of the path signature. Elements in the image of the signature map are represented by a formal power series
\begin{align}
\begin{aligned}
    \Sig(\phi) = \sum_{k=0}^{\infty} \sum_{ i_1,i_2,\dots,i_k} \Sig_{i_1,\dots,i_k}(\psi) e_{i_1} \otimes \cdots \otimes e_{i_k}
\end{aligned}
    \label{eq:formal-power-series}
\end{align}
Hence, as a feature map, the path signature transforms a path into a (infinite-dimensional) vector space of features. As a practical consideration, paths can be represented by a finite number of features via the \emph{truncated path signature} $\Sig(\psi)^{\leq p}$ \emph{at depth} $p$, obtained from \eqref{eq:formal-power-series} where the limit of the outer summation is truncated at $k = p$, i.e. $\Sig_{i_1,\dots,i_k}$ is set to zero for multi-indices with $k>p$. The number of features of the truncated path signature is given by $(n^{p+1} - 1)/(n - 1) - 1$; if the path is interpolated from a data stream, the number of extracted features, then, is independent of the length of the stream. In practice, this is convenient because you can compare datastreams with different lengths (i.e.~number of samples) in the same euclidean space.  The number of features in the truncated path signature grows rapidly. For example, for path with $n=3$ dimensions, there are $39$ features at a depth $3$, and $120$ features at a depth $4$, and $363$ features at a depth $5$.

\subsubsection{Properties of path signatures}
\label{sec:path-properties}

What do signatures actually measure? Are they interpretable? We recall some facts about signatures which make them appealing for feature engineering.
\paragraph{Compositionality \cite{chen1957}.}%
If $\psi$ and $\psi'$ are two paths, then $\Sig(\psi \ast \psi') = \Sig( \psi) \otimes \Sig( \phi')$ where $\ast$ is path concatenation and $\otimes$ is the multiplication operation inherited from the power series representation \eqref{eq:formal-power-series}. As a practical implication, if new data arrives, the signature does not need to be recomputed ``from scratch.''
\paragraph{Reparameterization invariance \cite{lyons1998}.}%
Suppose $\eta: [0,T] \to [0,T]$ is continuously differentiable, surjective and increasing, and $\psi \in \Path_{0,T}(\R^d)$. Then, $\Sig(\psi \circ \eta) = \Sig(\psi)$. Thus, the signature captures the sequential nature of a datastream, agnostic of the duration between samples. This property proves helpful when some GPS stations experience outages.
\paragraph{Uniqueness \cite{hambly2010}.}%
If $\psi$ is continuous and piece-wise smooth, then $\psi$ is uniquely determined by $\Sig(\hat{\psi})$ up to translation. In fact, it suffices that at least one coordinate of the path is monotone \cite{fermanian2021}, and a path signature can be ``inverted'' to produce a path which approximates the original path \cite{fermanian2023}.
\paragraph{Universality \cite{arribas2018}.}%
Any real-valued continous function on the space of paths $\Path_{0,T}\left(\R^d\right)$ can be approximated arbitrarily well by a linear functional on $T((\R^d))$. This suggests that path signatures form linearly-discriminant features.    

\subsubsection{Spatio-temporal path signatures}

Suppose $\G = (\V,\E)$ is an undirected graph. A \emph{spatio-temporal path} on a graph $\G$ is a map $\bpsi: [0,T] \to \R^{N \times n}$
specifying a path $\bpsi_i: [0,T] \to \R^n$ for each node $i \in \V$. Paths can be obtained from a spatial-temporal path on $\G$ by either restricting the path to a particular node or a particular feature. As a matter of notation, we use subscripts for the path $\bpsi_i$ at $i \in \V$ and superscripts for the path $\bpsi^f$ restricted to feature $f$; then, $\psi_i^f: [0,T] \to \R$ is simply the timeseries for feature $f$ at node $i$. Spatio-temporal paths are generated from datastreams indexed by $\V$, e.g.~from readings of sensors in a sensor network. In our application to monitoring slow slip events, the nodes are stations with GPS sensors and the datastreams are measurements of displacement in the east/west, north/south and up/down directions. Given a spatio-temporal path $\bpsi$, it's truncated \emph{spatio-temporal path signature} is defined as a matrix $\Sig^{\leq p}(\bpsi) \in \R^{N \times F}$ where $F = n(n^{p} - 1)/(n - 1)$; the rows of $\Sig^{\leq p}(\bpsi)$ are precisely the $\Sig^{\leq p}(\bpsi_i)$.

\subsection{Graph convolutional neural networks (GCNNs)}

Suppose $\G = (\V,\E)$ is an undirected graph with $N$ nodes $\V = \{1,2\dots,N\}$ and $M$ edges $\E \subseteq \V \times \V$. Graph neural networks (GNNs) are architectures for processing data assigned to the nodes and/or edges of $\G$. Graph convolutional neural networks (GCNNs) utilize a notion of convolution of graph signals by filters, introduced in the graph signal processing (GSP) literature \cite{ortega2018}. A \emph{graph signal} is a real-valued function on $\V$, often written as a column vector $\bx \in \R^N$. In GSP, signals are filtered by applying a graph shift operator (GSO), a matrix $\bS \in \R^{N \times N}$ whose sparsity pattern reflects the topology of the graph. Examples of GSOs include the (normalized) adjacency matrix, (normalized) Laplacian matrix. These GSOs and many others are formed from the adjacency matrix $\bA$ and the degree matrix $\bD$ of the graph.
Applying a GSO to a signal $k$ times, $\bS^k \bx$, aggregates local information in the $k$-hop neighborhood of each node in $\V$. Graph filters apply iterated graph shifts in order to select information at various spatial scales. Suppose $\bx \in \R^N$ is a graph signal, then, the output of a (linear shift-invariant) \emph{graph filter} is written
\begin{align*}
    \bH(\bS) \bx := \sum_{k=0}^{\infty} h_k \bS^k\bx
\end{align*}
where $\bh = \{h_k\}_{k=0}^{\infty}$ is a sequence of filter coeficients.

Suppose $\G$ has node features $f \in \{1,\dots,F\}$, and suppose $\bX \in \R^{N \times F}$ is a feature matrix on $\G$. A \emph{convolutional graph neural network} (GCNN) is a cascade of graph filters followed by point-wise nonlinear activation functions, represented by as sequence of layers $\ell \in \{ 0, 1, 2 \dots, L\}$, each with $F_{\ell}$ features. Each layer $\ell$ consists of an input $\bX_{\ell-1} \in \R^{N \times F_{\ell-1}}$ and an output $\bX_{\ell} \in \R^{N \times F_{\ell}}$,
\begin{align}
    \bX_{\ell} = \sigma_{\ell} \bigl( \sum_{k=0}^{K} \bS^k \bX_{\ell-1} \bH_{\ell k}  \bigr) \label{eq:gcnn}
\end{align}
where $\bH_{\ell k} \in \R^{F_{\ell} \times F_{\ell-1}}$ is the matrix of learned weights. Composing layers, we write $\bY = \Phi(\bS,\bX)$ for the entire GCNN. Among other types of GNNs, GCNNs such as ChebNet \cite{defferrard2016} and GCN \cite{kipf2016} have advantageous properties of permutation equivariance as well as stability to perturbations \cite{gama2020}.








\subsection{Path signature to GNN pipeline}
\label{sec:pipeline}

As our main contribution to methodology, we propose a machine learning pipeline, which we call PS-GCNN, to perform tasks on irregular sequential data collected by sensors in different locations in the spatial domain. Although we focus on graphs constructed by thresholding pairwise distances between locations, e.g.~sensor networks, our pipeline, summarized in Figure \ref{fig:summary}, can be applied to spatio-temporal paths on arbitrary graphs.

\subsubsection{Node feature engineering}

The input to our pipeline is set of locations $\{ \by_i \}_{i=1}^N \subseteq \mathcal{Y}$ in a metric space $(\mathcal{Y},d)$ as well as a collection of datastreams $\{ \mathcal{D} _i \}_{i=1}^N$ of $n$-dimensional data collected at each point in the metric space, i.e.~by a sensor. In the synthetic experiment, the metric space is $[0,1]^2$ with Euclidean distance; in the experiment with real data, a geographic region with geographic distance. By the reparameterization property (Section \ref{sec:path-properties}), the lengths of the datastreams may vary from node to node while still producing the same number of node features. Each sensor $i \in \{1,2,\dots,N\}$ receives several signals, each of which is viewed as a component of a path $\bpsi_i$, constructed by linear interpolation. The paths at each sensor, together, form a spatio-temporal path $\bpsi$. For a given depth, the truncated signature of the spatio-temporal path $\bpsi$ is computed to produce node features $\bX_{0} = \Sig^{\leq p}(\bpsi) \in \R^{N \times F}$. We use the open-source library \emph{Signatory} to compute path signatures \cite{kidger2020}.

\subsubsection{Proximity graphs}
Given $\{ \by_i \}_{i=1}^N \subseteq \mathcal{Y}$ and $\rho  \geq 0$ (hyperparameter), the $\rho$-\emph{proximity graph} is the graph $\G_{\rho} = (\V,\E_{\leq \rho})$ with nodes $\V = \{1,2,\dots,N\} $ and edges $\E_{\leq \rho} = \{ (i,j) \in \V^2: d(\by_i, \by_j) \leq \rho \}$. Proximity graphs are ubiquitous in machine learning, particularly when capturing the geometry of a manifold from data \cite{belkin2003,singer2012}.
The node features, then, are the input layer of a convolutional graph neural network (GCNN) on $\G_{\rho}$. In Section \ref{sec:experiments}, we observe the effect of $\rho$ on the performance of our model on a node regression and node classification task. 

\subsubsection{GCNN architecture}
\label{sec:architecture}

The graph module of our model is based on GCN \cite{kipf2016}, which is a GCNN with $\bH_{\ell k} = \mathbf{0}$ for $k \neq 1$ and graph shift operator $\bS = \tilde{\bD}^{-1/2} \tilde{\bA} \tilde{\bD}^{-1/2}$ where $\tilde{\bA} = \bA + \bI$ for $\bA$, the adjacency matrix of $\G$, and where $\tilde{\bD} = \bD + \bI$ for $\bD$, the degree matrix of $\G$. If $\G$ is completely disconnected, for this choice of graph shift operator $\bS = \bI$, which implies the resulting GCNN is reduced to a multi-layer perceptron (MLP). As a consequence, testing our model on $\G_{\rho}$ for $\rho = 0$ is equivalent to testing it on an MLP.

\begin{figure}[h!]
    \centering
    \includegraphics[width=0.75\linewidth]{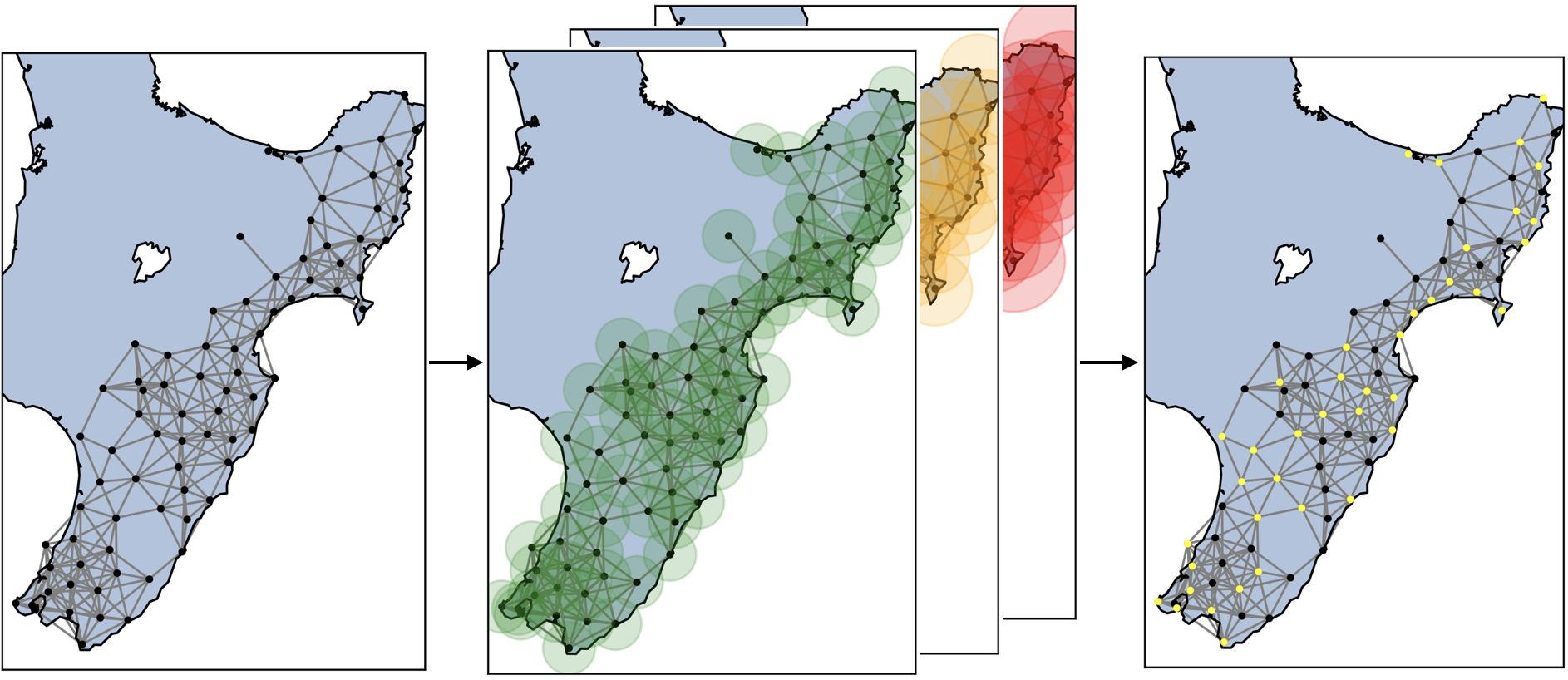}
    \caption{Graph convolutional neural network (GCNN) node classifier with three convolutional layers. Outputs are binary labels which convey membership in a hypothesis class (yellow or black).}
    \label{fig:gnn}
\end{figure}

In all of our experiments, our GNN model has $L=3$ layers $\ell \in \{0,1,2,3\}$. $F_0$ is the number of input features, determined by the number of dimensions of the time series and the truncated signature depth in the Path Signatures module. The hidden layers $\ell = 1, 2$ are chosen to have $F_1 = 32$ and $F_2 = 16$ features. We use a ReLU activation function $\sigma_\ell(x) = \max(0,x)$ for $\ell = 0, 1, 2$, and either the identity or $\sigma_{\ell}(x) = \mathrm{sigmoid}(x)$ for $\ell = L$, depending on whether the task is a classificaiton or a regression. The GNN module of our pipeline is summarized in Figure \ref{fig:gnn}. We briefly recall the particular difference between our regressor and classifier architectures.

\begin{itemize}
    \item[] \textit{Classifier GCNN Module.} The output of the regressor $F_L$ is equal to the number of parameters the model is trying to predict. The readout activation function is a sigmoid, and the loss is mean-squared error. We use the GCNN regressor for our experiments on simulated data (Section \ref{sec:simulated}).
    \item[] \textit{Regressor GCNN Module.} The output of the classifier is two features $F_L = 2$. There is no readout readout activation, and the loss is binary cross-entropy. We use the GCNN classifier module for our experiments on real data (Section \ref{sec:real-data}).
\end{itemize}

\section{Experiments}
\label{sec:experiments}

We demonstrate the efficacy of using path signatures as node features in graph neural networks (GNN) for both regression and classification tasks.  We report the results of several experiments on both simulated and real data. In both type of experiment, we generate proximity graphs, in the simulated experiments from uniformly sampled points in the interior square, in the real-work experiments from the locations of sensors tracking movement of the Earth's crust.

In the simulated experiment, we train a GCNN regressor in order to estimate the parameters of a discrete reaction diffusion equation \eqref{eq:reaction-diffusion} for two different types of reactions, the Lotka-Volterra equations, modeling the populations of predators versus prey, and the FitzHugh Nagumo, modeling excitable systems such as the neuron. In the experiments on real data, we train a GCNN classifier to predict the labels of GPS time series signals as behaving as either a linear or nonlinear system.

\subsection{Simulated experiments: reaction-diffusion}
\label{sec:simulated}

The nonlinearities of the signal are typically attributed to positive maximum Lyapunov exponents \cite{kantz2004nonlinear} and thus the presence of a chaotic (nonlinear) system of equations underpinnng it. Reaction-diffusion partial differential equations (PDEs) are such nonlinear systems that are also shown to model the pressure and temperature of the subduction interface, leading to slow earthquake events \cite{poulet2013}. As a preliminary step, with the eventual goal to learn parameters from real data, we design a synthetic experiment involving recovering parameters of a stochastic reaction-diffusion PDE. Another purpose of the synthetic experiments is to establish a benchmark for PS-GCNN. 

\subsubsection{Disretization}

Our approach is to discretize the spatial coordinates of the reaction-diffusion equations \eqref{eq:reaction-diffusion} by uniformly sampling points in  the unit square $[0,1]^2$ and constructing a proximity graph with a fixed threshold radius $\rho > 0$, thus converting the PDE to an system of ordinary differential equations (ODEs) whose variables are indexed by the nodes of a graph. If $\rho = 0$, then the there is no coupling between the local reaction-diffusion dynamics; if $\rho = \sqrt{2}$, every pair of nodes is coupled. 

Thus, given $\G_{\rho}$ with adjacency matrix $\bA = [a_{i,j}]$, the stochastic reaction-diffusion equations for two reactants $u_i(t)$ and $v_i(t)$ over a graph nodes are given by the following system
\begin{align}
    \begin{aligned}
        d u_i &= \biggl[ R_u(u_i,v_i) - D_u \sum_{j = 1}^N a_{i,j}(u_i - u_j) \biggr] dt + \sigma_u d W_{t,i} \\
        d v_i &= \biggl[ R_v(u_i,v_i) - D_v \sum_{j =1}^N a_{i,j} (v_i - v_j) \biggr] dt  + \sigma_v d W_{t,i} \\
        i &= 1,\dots,N
    \end{aligned}\label{eq:discrete-reaction-diffusion}
\end{align}
where $D_u,D_y>0$ is the diffusivity, and $R_u$ and $R_v$ model the rate of reaction between $u_i$ and $v_i$, $W_{t,i}$ is a Weiner process, and $\sigma_u^2,\sigma_v^2$ are the variances of the process. To capture the dependence of $R_u$ and $R_v$ on the spatial domain, we suppose each node has reaction terms that depend on local parameters.

\subsubsection{Simulating the dynamics}

We generate a synthetic dataset based on the trajectories of \eqref{eq:discrete-reaction-diffusion} for $\G_{\rho}$. Note that $\rho$ is a parameter that effects the connectivity of $\G_{\rho}$, and, thus, along with $D_u$ and $D_v$, affects diffusion. We choose the Lotka-Volterra (LV) which model the evolution of predator versus prey populations \cite{lotka1910contribution} and the FitzHugh Nagumo (FHM) equations \cite{fitzhugh1995}, which model excitable systems such as the neuron. In the LV equations, the reaction terms of \eqref{eq:discrete-reaction-diffusion} are given by
\begin{align}
    \begin{aligned}
        R_u(u_i,v_i) &=& \alpha_i u_i - \beta_i u_i v_i \\
        R_v(u_i,v_i) &=& \delta_i u_i v_i - \gamma_i v_i
    \end{aligned} \quad i \in \V \label{eq:LV}
\end{align}
with node-variant parameters $\alpha_i,\beta_i,\delta_i,\gamma_i \geq 0$. Similarly, in the FHM equations, the reaction terms of \eqref{eq:discrete-reaction-diffusion} are given by
\begin{align}
    \begin{aligned}
        R_u(u_i,v_i) &=& u_i - u_i^3/3 - v_i \\
        R_v(u_i,v_i) &=& \epsilon_i (u_i + a_i - b_i v_i)
    \end{aligned} \quad i \in \V \label{eq:FHN}
\end{align}
with node-variant parameters $\epsilon_i, a_i, b_i \geq 0$. As a remark, in the case of the LV equations, the system \eqref{eq:discrete-reaction-diffusion} is a linear stochastic process for $\beta = \delta = 0$. For each reaction model, we chose initial conditions $\bu(0) \in \R^N$ and $\bv(0) \in \R^N$ randomly in the unit interval. We, then, apply the Euler method with $\Delta t = 0.01$ and time-horizon $T = 100$ to obtain trajectories, to be converted into spatio-temporal paths $\bpsi_i(t) = \left(u_i(t),v_i(t)\right)$, $i \in \V$. Trajectories for a node and several neighbors are plotted in the Appendix (see Fig.~\ref{fig:simulated-data}) to demonstrate how the signals of neighboring nodes interact over time.

\subsubsection{Results: effect of diffusion}

Then, we use our pipeline (see Fig.~\ref{fig:summary}) to process the resulting spatio-temporal paths for a regressor GCNN module (Section \ref{sec:architecture}). The regression task we design is to predict the parameters \eqref{eq:discrete-reaction-diffusion} for each node, both in the case of the Lotka-Voterra (LV) equations \eqref{eq:LV} and the FitzHugh Nakumo equations \eqref{eq:FHN}. In the LV equations, the output of the GCNN, then, has $F_L = 4$ features because there are $4$ parameters for each node, while in the FHN equation, $F_L = 3$. To observe the effect of diffusion on the accuracy of our predictions, we tested our model for various levels diffusivity, assuming that $D = D_u = D_v$ for simplicity. The results are reported in Table \ref{tab:diffusion}.

\begin{table}[h!]
  \centering
  \small
  \begin{tabular}{lccc}
    \toprule
     & \multicolumn{2}{c}{MSE} \\
    \cmidrule(r){2-3}
     Diffusivity ($D$)   & LV & FHN \\
    \midrule
    $0.01$ & $0.1692 \pm 0.1144$ & $0.1003 \pm 0.0668$ \\
    $0.02$ & $0.2081 \pm 0.1257$ & $0.0923 \pm 0.0500$ \\
    $0.03$ & $0.2263 \pm 0.1146$ & $0.0849 \pm 0.0087$ \\
    $0.04$ & $0.2403 \pm 0.1127$ & $0.0864 \pm 0.0096$ \\
    $0.05$ & $0.1841 \pm 0.1098$ & $0.0861 \pm 0.0108$ \\
    $0.06$ & $0.1019 \pm 0.0702$ & $0.0835 \pm 0.0060$ \\
    $0.07$ & $0.0873 \pm 0.0278$ & $0.0828 \pm 0.0043$ \\
    $0.08$ & $0.1044 \pm 0.0550$ & $0.0839 \pm 0.0064$ \\
    $0.09$ & $0.0821 \pm 0.0051$ & $0.0846 \pm 0.0063$ \\
    $0.10$ & $0.0813 \pm 0.0055$ & $0.0843 \pm 0.0053$ \\
    \bottomrule
  \end{tabular}
  \caption{Accuracy of our predictions of the parameters of the Lotka-Volterra (LV) and FitzHugh-Nagumo (FHN) models for several diffusivity constants ($D$). Mean-square-error (MSE) is reported as an average across $10$ trials and $4$ folds.}
  \label{tab:diffusion}
\end{table}

We observed for the Lotka-Voterra simulation, that, when diffusivity is low, the performance of the model suffers. The performance of the regressor in the FHN simulation appears to be less sensitive to diffusivity. At least in the case of the LV simulation, our results suggest that the graph neural network is learning not only the reaction coefficients for every node but also the overall diffusion process.


\subsection{Experiments on real data: GPS time series}
\label{sec:real-data} 

We now demonstrate how to apply our methods to real data where a physical model of a system is unknown, or, in the case of slow slip events, we cannot recover the true state space of the system from noisy observables, e.g.~displacement \cite{truttmann2023}. We apply PS-GCNN, here, on the task of classifying nonlinear features appearing in different locations, a node-level binary classification task on a $\rho$-proximity graph representing the network. In these experiments, we treat $\rho$ as a hyperparameter because the ``true'' radius of interaction between signals is unknown.

\subsubsection{Node labels}

GeoNet Aotearoa New Zealand Continuous GNSS Network (\url{https://fits.geonet.org.nz}), abbreviated GeoNET, is a GPS sensor network collecting data throughout New Zealand (see Fig.~\ref{fig:summary}). The sensors take daily readings of the the displacement in millimeters (mm) in three directions: east/west, north/south and up/down. We restrict our attention to a hand-picked subset of this network consisting of $80$ of those sensors located in the east coast of the north island. Our task is to make predictions about the signals observed by each station, assigning them to one of two classes: \emph{linear} or \emph{nonlinear}.

A commonly used technique for examining nonlinearity within a time series derived from an unknown dynamical system is surrogate data hypothesis testing \cite{lancaster2018,kantz2004nonlinear}. The null hypothesis, in this test, is that the displacement time series was generated by a linear stochastic process. (Note that the rejection of the null hypothesis does not definitively rule out with  the possibility of nonlinearity in these cases.) In previous work \cite{truttmann2023}, this method was successfully applied, allowing for the dismissal of the null hypothesis in $11$ different GPS time series, which we assign to the class \emph{nonlinear}. The remaining $53$ labeled stations, for which the null hypothesis could not be rejected, are assigned to the class \emph{linear} (see Fig.~\ref{fig:ground-truth} in Appendix). Note that class imbalances are treated by weighting the cross-entropy loss.


\subsubsection{Results: tuning the radius}

\begin{figure}[h]
    \centering
        \includegraphics[width=0.45\linewidth]{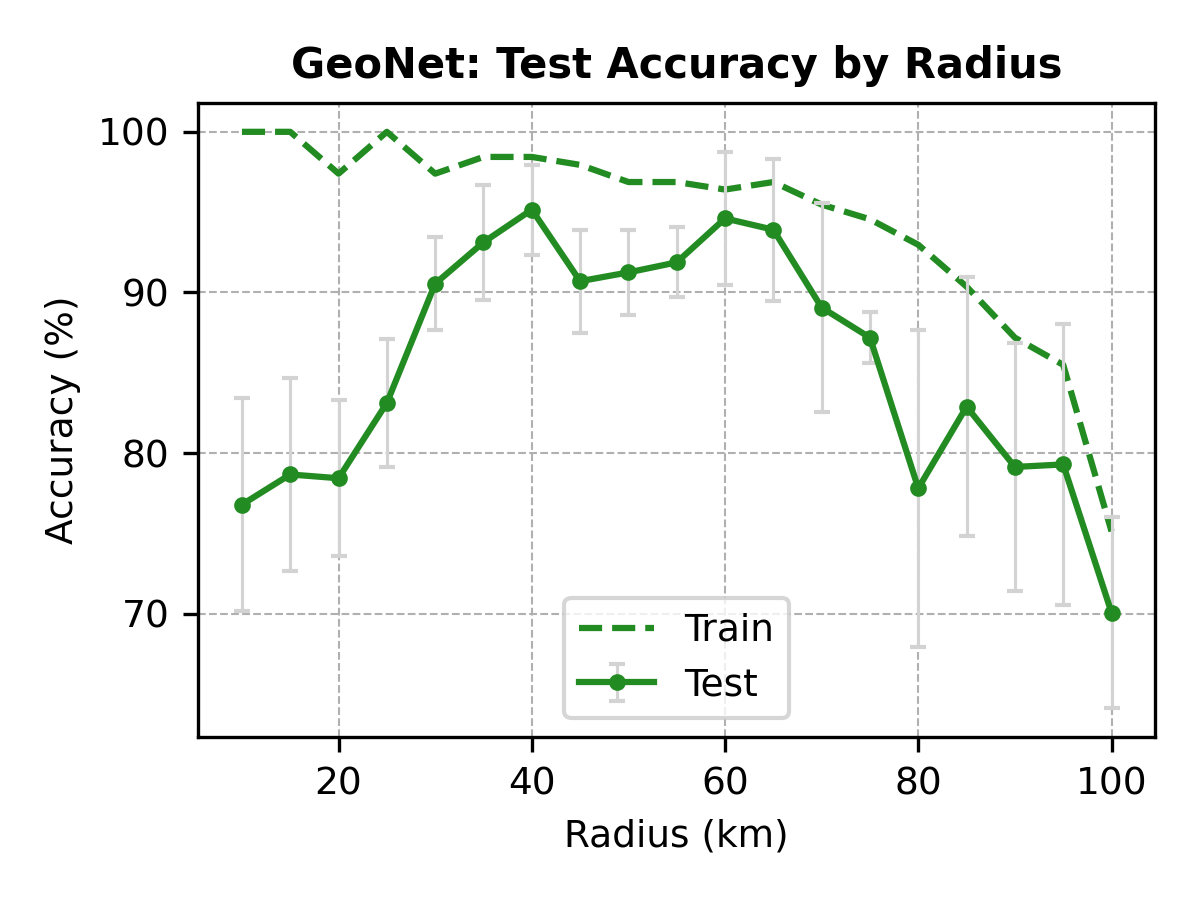}
        \quad
        \includegraphics[width=0.45\linewidth]{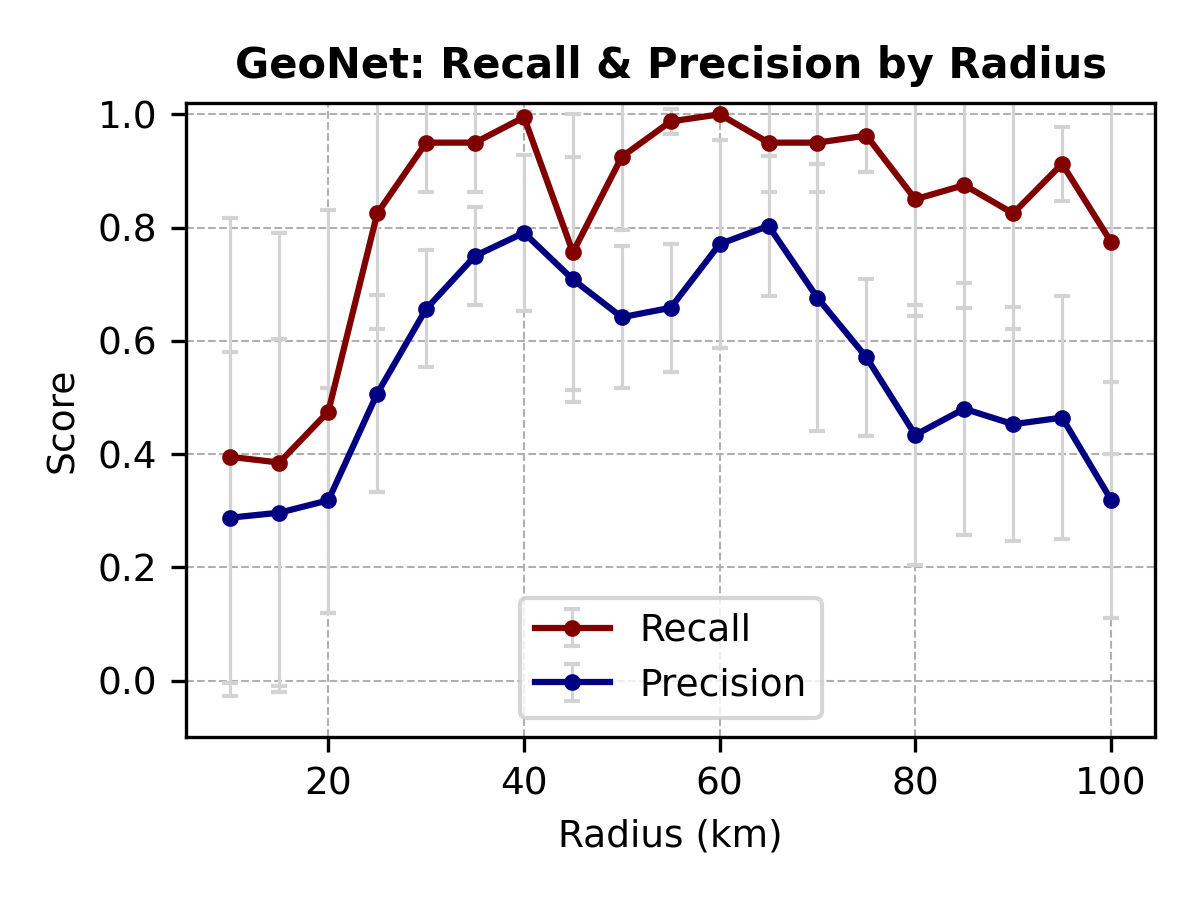}
        \caption{Tuning the radius hyperparameter $\rho$.}
    \label{fig:radius}
\end{figure}

As before, we construct a $\rho$-proximity graph representation of the data in the spatial domain. The main difference, now, is that the locations of the sensors are not sampled uniformly and the threshold radius needs to be tuned. In the process of tuning, we analyze the accuracy, precision, and recall of the model as a function of $\rho$ (see Fig.~\ref{fig:radius}).
We find that the optimal threshold radius is around $40$ or $60$ km. We comment that models with a small radius ($\leq 10$ km) are equivalent to multi-layer perceptions (MLPs) (with signature features), obtaining around $70$-$80 \%$ accuracy. The remaining improvement in the model, thus, is the result of the spatial structure of the signals, learned by graph filters. We also note that a large threshold radius results in significantly degraded precision with marginally worse recall. In this case, the classifier will have the tendency to reject the null hypothesis.

\subsection{Ablation study: signatures v.s.~summary statistics}

Finally, we perform an ablation study comparing our path signature based- GCNN method with an identical GCNN, instead with summary statistics of the timeseries as node features. Our method and the ablation are tested on both of simulated datasets and the real dataset. The first two datasets are trajectories of the Lotka-Voterra (LV) and FitzHugh Nagumo (FHN) stochastic discrete reaction-diffusion equations \eqref{eq:discrete-reaction-diffusion} with diffusivity $D_u = D_V = 0.05$, noise intensity $\sigma_u = \sigma_v = 10$, over a $\rho$-proximity graph with $100$ nodes and $519$ edges with $\rho = 0.2$. The regressor GCNN, for the simulated data, is trained with Adam under mean-squared-error loss with learning rate $0.01$ and weight decay $0.005$. The real-world dataset, GeoNET, consists of the same GPS time series and stations as before. For the semi-supervised node classification task, the classifier GCNN module is trained with Adam under mean-squared-error loss with learning rate $0.001$ and weight decay $0.005$.  In all experiments, path signatures are truncated at a depth of $4$, and summary statistics are generated by an off-the-shelf feature-extraction package (see \citet{christ2018} for details). The results are reported in Table \ref{tab:ablation}.

\begin{table}[h]
\centering
\small
\begin{tabular}{lcccccc}
\toprule
& \multicolumn{2}{c}{Lotka-Voterra (LV)} & \multicolumn{2}{c}{FitzHugh Nagumo (FHN)} & \multicolumn{2}{c}{GeoNET} \\
\cmidrule(r){2-3} \cmidrule(r){4-5} \cmidrule(r){6-7}
Node Feature Method & MSE & MAE & MSE & MAE & Accuracy & F1 \\
\midrule
PS-GCNN (ours)  & $0.08 \pm 0.03$ & $0.25 \pm 0.03$ & $0.08 \pm 0.0046$ & $0.24 \pm 0.0094$ & $96.41 \% \pm 4.61\%$ & $0.92 \pm 0.10$ \\
SS-GCNN (ablation) & $0.25 \pm 0.11$ & $0.42 \pm 0.11$ & $0.24 \pm 0.12$ & $0.41 \pm 0.12$ & $92.03 \% \pm 5.41\%$ & $0.76 \pm 0.18$ \\
\bottomrule
\end{tabular}
\caption{Ablation study comparing our method to the same architecture, instead with summary statistic node features (SS-GCNN). In the simulated experiments, mean-squared-error (MSE) and mean-absolute-error (MAE) are reported over $10$ trials and $4$ folds. In the experiments with real data, the accuracy and F1 scores are reported over $10$ trials and $4$ folds.}
\label{tab:ablation}
\end{table}

\section{Related Work}
\label{sec:related}

The intersection of physics and machine learning has been a recent hive of activity, especially the general program to integrate physics-based models and machine learning techniques \cite{willard2020}. Neural networks have been utilized to simulate complex physical systems, some with GNNs, including deformations of materials \cite{sanchezgonzalez2020}, many others with multi-layer perceptions (MLP), including environmental flow fields \cite{khodayimehr2023,khodayimehr2020,khodayimehr2020b}. Graph machine learning with time series for tasks such as forecasting, anomaly detection, time series classification, and imputation has received considerable attention \cite{jin2023}. There are several recent GNN architectures tailored to processing time-variant graph signals \cite{cini2023}. Graph recurrent neural networks (GRNS) utilize a LSTM alongside graph filters \cite{ruiz2020}. Space-time graph neural networks (ST-GNNs) utilize a notion of diffusion that mimics the diffusion process of time-variant graph signals \cite{hadou2021}. The sequence to path signature feature to neural network pipeline was first introduced in \citet{kidger2019}. Path signatures have been previously used as spatio-temporal representations for skeleton-based human action identification  \cite{yang2022}. They compute signatures for the motions of each joint (represented as a node of a graph) and, then, concatenate all the features as an input to a single-layer fully-connected neural network (FCNN). Our approach is different, as we use the path signature features as node features for a graph neural network. We, too, utilize a shallow network, but we exploit graph convolutional layers to capture the propogation and mixture of the spatio-temporal signals.

\section{Conclusion}
\label{sec:conclusion}

After motivating why signatures and graph neural networks (GNNs) are appropriate for the analysis of GPS time series, we introduced a path signature to graph convolutional neural network pipeline (PS-GCNN), illustrated in Figure \ref{fig:summary}. We tested our method on preliminary tasks: a regression task for data simulated by stochastic reaction-diffusion PDEs, and a classification task for GPS time series measuring land surface displacement for the prediction of slow slip events. In the future, we want to use our framework to inform GPS sensor placement, inferring an optimal ``radius of interaction'' (see Fig.~\ref{fig:radius}), as well as triangulation of slow earthquake signals. To overcome the challenge of a lack of labeled nodes,  we plan to use self-supervised learning \cite{xie2023}, along with signature methods, to make inferences about slow slip events at scale.

\section*{Broader Impact}
\label{sec:impact}
The method developed here can have broad implications in the societal response of seismic hazard, potentially allowing for both a better understanding of the distribution of earthquakes in an area, but also opening the door to forecasting and triangulation of the slip signal of earthquakes. Furthermore, the method suggested here can be deployed on any network of sensors capturing irregular (noisy, chaotic or stochastic) time-series data and tasked with various predictive tasks including classification, anomaly detection, source localization, forecasting and so forth. Indicative applications to which this method can be extended on include network neuroscience \cite{bassett2017} and environmental monitoring.

\bibliography{paper}

\newpage
\appendix
\onecolumn

\section{Additional figures}

\begin{figure}[h]
    \centering
    \includegraphics[width=0.75\linewidth]{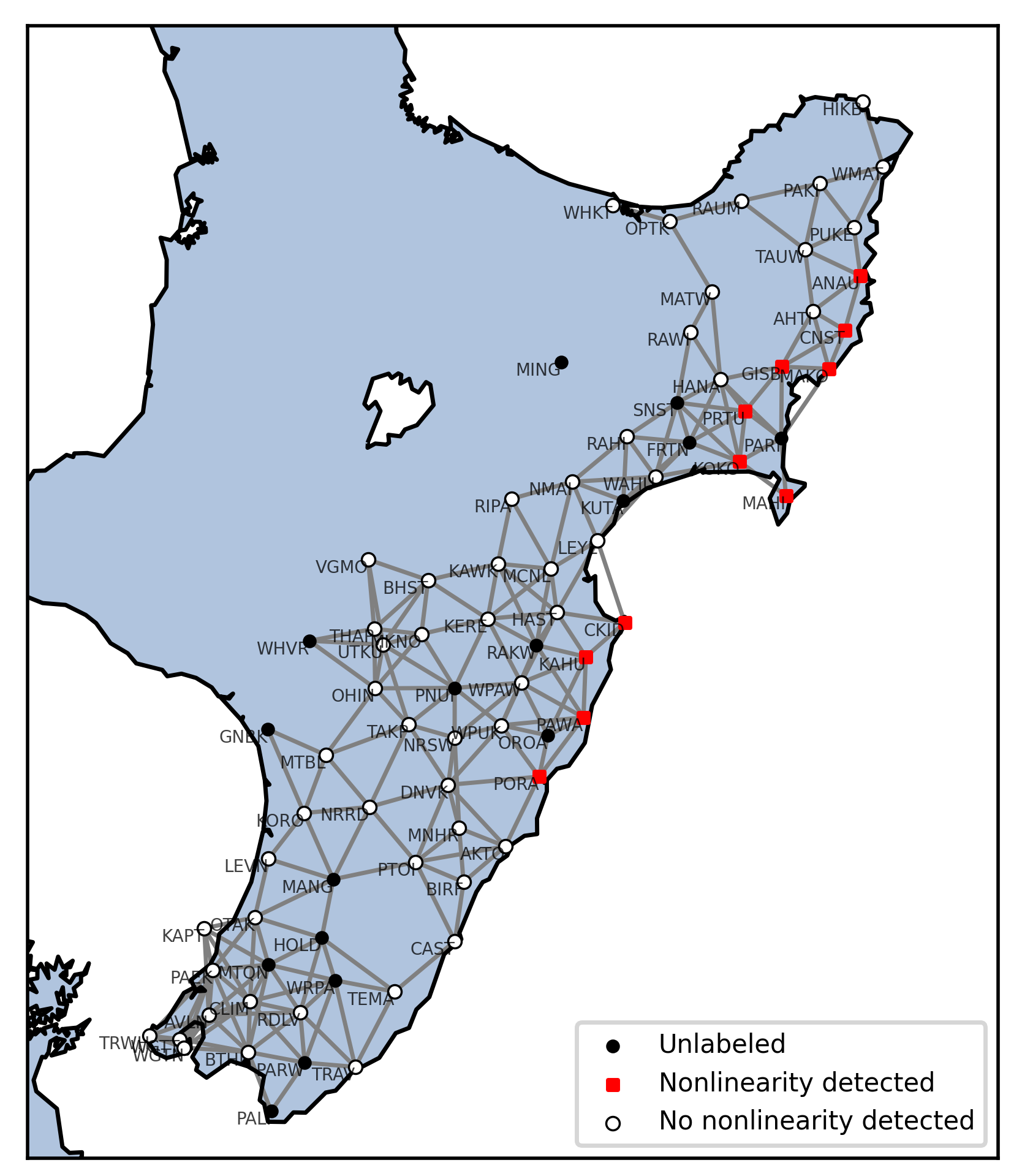}
    \caption{Ground truth labels for our semi-supervised node classification task. Of the $80$ stations considered, $64$ are labeled; nonlinearity is detected in $11$ of them, while the remaining $53$ stations had no nonlinearity detected. The proximity graph pictured has a threshold radius $\rho = 40$ km.}
    \label{fig:ground-truth}
\end{figure}

\begin{figure}[h!]
    \centering
    \includegraphics[width=0.75\linewidth]{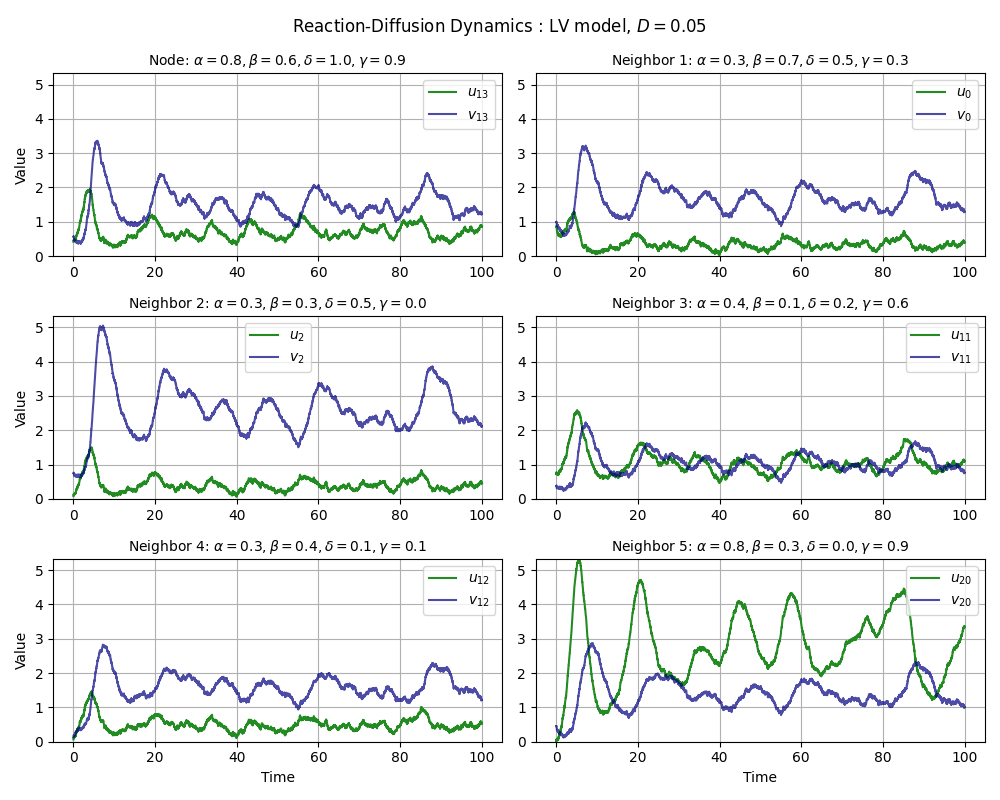}
    \caption{A comparison between trajectories of the Lotka-Volterra equations (above) and the FitzHugh Nagumo equations (below) for a randomly chosen node and a few of its neighbors.}
    \includegraphics[width=0.75\linewidth]{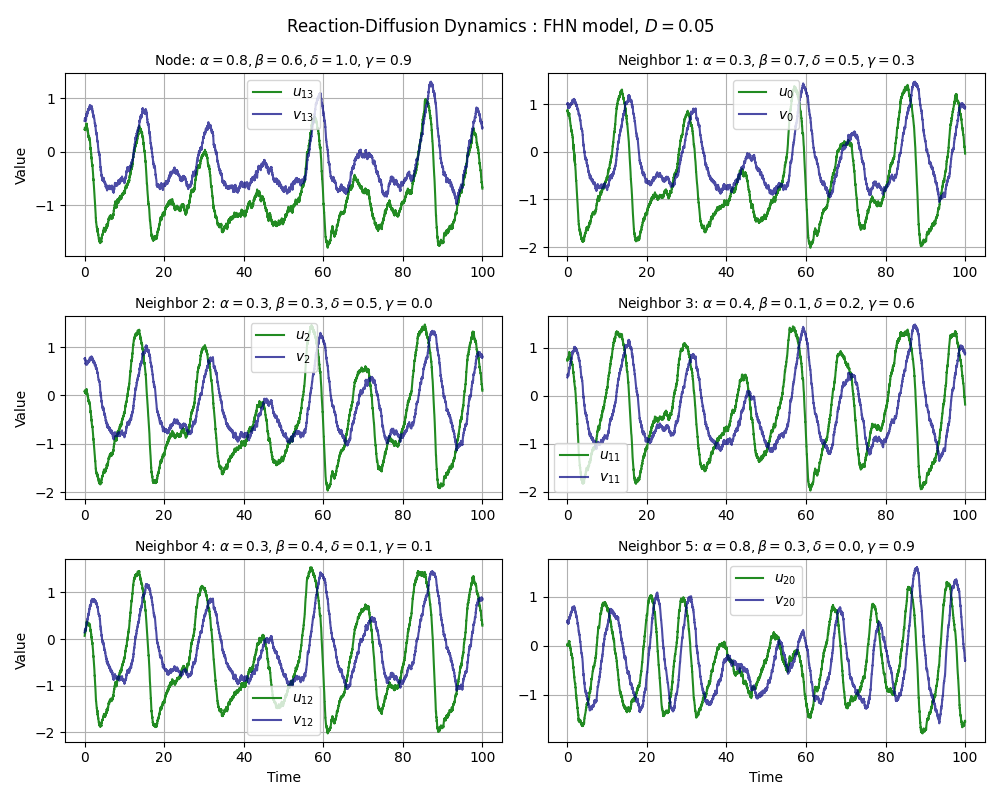}
    \label{fig:simulated-data}
\end{figure}

\end{document}

%% file: macros.tex
\renewcommand{\phi}{\varphi}
\renewcommand{\leq}{\leqslant}
\renewcommand{\geq}{\geqslant}

\newcommand{\bx}{\mathbf{x}}
\newcommand{\by}{\mathbf{y}}

\newcommand{\bh}{\mathbf{h}}
\newcommand{\bX}{\mathbf{X}}
\newcommand{\bY}{\mathbf{Y}}
\newcommand{\bu}{\mathbf{u}}
\newcommand{\bv}{\mathbf{v}}

\newcommand{\bA}{\mathbf{A}}

\newcommand{\bD}{\mathbf{D}}
\newcommand{\bS}{\mathbf{S}}
\newcommand{\bH}{\mathbf{H}}
\newcommand{\bI}{\mathbf{I}}

\newcommand{\bpsi}{\boldsymbol{\psi}}

\newcommand{\G}{\mathcal{G}}
\newcommand{\V}{\mathcal{V}}
\newcommand{\E}{\mathcal{E}}

\newcommand{\R}{\mathbb{R}}

\DeclareMathOperator{\Sig}{Sig}

\DeclareMathOperator{\Path}{Path}